\documentclass{article}
\usepackage{spconf,amsmath,amssymb,graphicx}
\usepackage[T1]{fontenc}
\usepackage{booktabs}
\usepackage{cite}
\usepackage{capt-of}
\usepackage{orcidlink}
\usepackage{microtype}
\usepackage{xcolor}
\definecolor{BLACK}{named}{black}
\usepackage{tikz}
\usepackage{hyperref}

\colorlet{RED}{red}

\hypersetup{
pdftitle={Handwritten Digit Leakage from Smartphone Motion Sensors Across Unseen Users and Phone Models},
pdfauthor={Alvarez Casado et al.},
pdfsubject={Motion side channel evaluation on HuMIdb}
}

\newcommand{\nacc}{\mathbf{a}}
\newcommand{\lacc}{\boldsymbol{\ell}}
\newcommand{\gyro}{\boldsymbol{\omega}}
\newcommand{\X}{\mathbf{X}}
\DeclareMathOperator*{\argmax}{arg\,max}
\DeclareMathOperator{\clamp}{clamp}

\definecolor{splittr}{HTML}{0072B2}
\definecolor{splitva}{HTML}{BDBDBD}
\definecolor{splitte}{HTML}{E69F00}

\makeatletter
\long\def\@makecaption#1#2{
 \vskip 4pt
 \setbox\@tempboxa\hbox{#1. #2}
 \ifdim \wd\@tempboxa >\hsize #1. #2\par \else \hbox
to\hsize{\hfil\box\@tempboxa\hfil}
 \fi}
\makeatother

\makeatletter
\def\@ssect#1#2#3#4#5{\@tempskipa #3\relax
  \ifdim \@tempskipa>\z@
    \begingroup \bf\centering \interlinepenalty \@M \uppercase{#5}\par\endgroup
  \else
    \def\@svsechd{#4{\hskip #1\relax #5}}%
  \fi
  \@xsect{#3}}
\makeatother

\newcommand{\teaser}{%
\begin{minipage}{\textwidth}
\centering\normalsize
\includegraphics[width=0.92\textwidth]{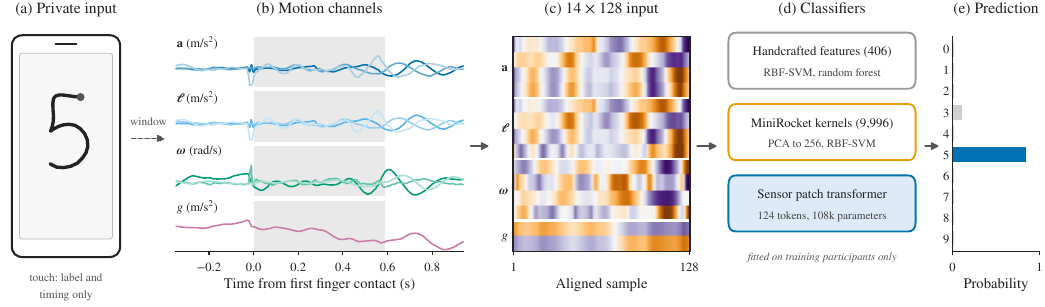}
\vspace{-2.5mm}
\captionof{figure}{\textcolor{black}{Overall view of the evaluation pipeline, illustrated with a correctly classified test recording of the digit 5.} (a) Finger trajectory, dot at first contact. Touch supplies only the label and the drawing interval. (b) Raw motion after mean removal, $x$, $y$, $z$ from dark to light, drawing interval in gray. (c) Aligned 14 channel input, standardised per channel for display. (d) Classifier families, fitted on training participants. (e) Frozen transformer probabilities.}
\label{fig:overview}
\end{minipage}}
\makeatletter
\def\@maketitle{\newpage
 \null
 \vskip 2em \begin{center}
 {\large \bf \@title \par} \vskip 1.5em {\large \lineskip .5em
\begin{tabular}[t]{c}\@name \\ \@address
 \end{tabular}\par} \end{center}
 \par
 \vskip 0.3em
 \teaser
 \vskip 0.6em}
\makeatother

\title{Handwritten Digit Leakage from Smartphone Motion Sensors \\ Across Unseen Users and Phone Models}

\name{\begin{tabular}{c} Constantino {\'A}lvarez Casado$^{\dagger}$$^{\ddagger}$$^{\orcidlink{0000-0002-3052-4759}}$, Erkka Rantahalvari$^{\ddagger}$$^{\orcidlink{0009-0004-2749-6592}}$, Matteo Pedone$^{\dagger}$$^{\orcidlink{0000-0003-4570-7264}}$, Matti Matilainen$^{\dagger}$$^{\orcidlink{0000-xxxx-xxxx-xxxx}}$, \\Manuel Lage Cañellas$^{\dagger}$$^{\orcidlink{0000-0002-4917-340X}}$, Le Nguyen$^{\dagger}$$^{\orcidlink{0000-0001-7765-1483}}$, Simo Hosio$^{\dagger}$$^{\star}$$^{\orcidlink{0000-0002-9609-0965}}$, Olli Silv\'en$^{\dagger}$$^{\ddagger}$$^{\orcidlink{0000-0002-2661-804X}}$, Miguel Bordallo L{\'o}pez$^{\dagger}$$^{\orcidlink{0000-0002-5707-9085}}$
\end{tabular}}

\address{
$^{\dagger}$University of Oulu, Oulu, Finland \\
$^{\ddagger}$Candour Ltd, Oulu, Finland \\
$^{\star}$University of Tokyo, Tokyo, Japan
}

\begin{document}
\ninept
\maketitle

{\let\thefootnote\relax\footnotetext{\textcolor{black}{This work has been submitted to the IEEE for possible publication. Copyright may be transferred without notice, after which this version may no longer be accessible. This extended version adds methodological details, a protocol schematic, implementation settings and further analyses to the conference submission.}}}

\begin{abstract}
Smartphone motion sensors support interactive applications, but their readings may also reveal touchscreen input beyond their intended use. Assuming known drawing intervals, we study whether handwritten digits remain predictable across users and devices, as a 10-class problem on 19,628 HuMIdb recordings from 481 participants. We compare handcrafted features with classical machine learning algorithms, MiniRocket kernels, and a compact sensor patch transformer on accelerometer, linear acceleration, gyroscope, and gravity signals. The transformer achieves 57.74\% accuracy and 82.64\% top-3 accuracy on 75 unseen participants, and 58.77$\pm$0.95\% over 3 seeds for unseen participants on 9 unseen phone models. Low motion recordings remain informative, accuracy is not monotonic in motion level, and the tested contrastive pretraining, augmentation, and derived signals give no consistent gains. Digits are thus predictable beyond familiar users and phone models under assumed segmentation, while acquisition-order shortcuts limit conclusions about practical privacy exposure. Code available at: \url{https://github.com/Arritmic/motion-digit-leakage}.
\end{abstract}


%
%
\section{Introduction}
\label{sec:introduction}

Smartphone motion sensors support applications that respond to device movement and user interaction\textcolor{black}{. However, measurements collected for these functions may also disclose information about touchscreen input. Finger contact and movement produce small accelerations and rotations of the phone, which previous studies have used to infer keystrokes, tap locations, passwords and handwritten patterns~\cite{cai2011touchlogger,xu2012taplogger,miluzzo2012tapprints,tang2018niffler,lee2018handwritten}.} An application or a recipient of recorded sensor data may thus infer content beyond the intended purpose of collection. \textcolor{black}{Handwritten numbers provide a concrete case for studying this risk, because they can represent personal or sensitive information, such as codes, amounts or dates, depending on the application. This indirect channel raises a privacy question that differs from authentication, namely how much of the entered content becomes predictable from motion alone.}

\textcolor{black}{Motion signals contain several sources of information that overlap. They reflect not only the interaction itself, but also how a person holds and moves the device. These behavioural characteristics support continuous authentication~\cite{sitova2016hmog,mahbub2019continuous}, wearable triaxial acceleration supports stress and affect detection~\cite{schmidt2018wesad}, smartphone recordings from walking and tapping tasks were collected at scale to quantify Parkinson disease symptoms~\cite{bot2016mpower}, and selfie capture motion supports identity verification~\cite{rantahalvari2026selfie}. Content inference asks a complementary question. Rather than identifying who performs an interaction, it seeks to determine what was entered despite differences between people and devices. This distinction matters, because successful recognition on familiar users or phones does not establish transfer to unfamiliar ones.}

\textcolor{black}{Two opposite expectations shape the interpretation of this question. The first is that the response of a phone to finger writing is too weak to survive changes of posture, grip and device. The second is that high accuracy is easy to obtain, because a model can memorise the person, the phone or the order of acquisition. Previous handwriting inference studies demonstrate feasibility under controlled recording conditions~\cite{lee2018handwritten}, which cannot separate these explanations, and generalization across participants and phone models under less constrained acquisition remains insufficiently characterized. The evaluation is further complicated by differences in motion level, sensor timing and recording protocols. A useful assessment must therefore examine both content predictability and the extent to which the evaluation supports conclusions about privacy exposure.}

\textcolor{black}{We investigate these questions through 10-class handwritten digit classification on HuMIdb~\cite{acien2021humidb}, where participants drew digits on their own phones without supervision, as shown in Figure~\ref{fig:overview}. We assume that the drawing intervals are known and use motion signals as the only input to the classifier. This formulation isolates recognition within a drawing interval, with the automatic detection of handwriting left outside the evaluated task. Each expectation is then tested directly. Partition protocols control whether the test people and phone models appear in training, a motion level analysis asks whether weak device motion prevents inference, and representation and training choices from prior work test whether larger models or invariance objectives improve transfer.} Our contributions are as follows:

\begin{enumerate}\setlength{\itemsep}{0pt}\setlength{\leftmargin}{1.3em}\setlength{\labelwidth}{1em}
    \item \textcolor{black}{An evaluation of three representation families using aligned multisensor signals, four partition protocols and uncertainty estimates clustered by participant or phone model, together with an examination of acquisition-order shortcuts.}
    \item \textcolor{black}{Evidence of digit predictability for unseen participants (57.74\% accuracy) and for unseen participants using unseen phone models (58.77\%). Low motion recordings remain informative, and the gyroscope is the strongest single sensor in the handcrafted feature comparison.}
    \item \textcolor{black}{An assessment of training and signal extensions showing no consistent accuracy gains from the tested contrastive pretraining, augmentation and derived channels, alongside analyses of participant overlap and training data selection.}
\end{enumerate}

%
%
\section{\texorpdfstring{\textcolor{black}{Related Work}}{Related Work}}
\label{sec:related}

\noindent\textbf{\textcolor{black}{Motion side channels.}} \textcolor{black}{TouchLogger, TapLogger and TapPrints demonstrated that onboard motion sensors can reveal keystrokes and touchscreen tap locations~\cite{cai2011touchlogger,xu2012taplogger,miluzzo2012tapprints}. Niffler used movements between consecutive taps to infer passwords across users~\cite{tang2018niffler}, and Lee et al.\ recognized handwritten patterns with dynamic time warping~\cite{lee2018handwritten}. Android sensor streams can also be sniffed and manipulated at the application level~\cite{mohamed2017smashed}, which widens the set of parties that may hold such recordings. These studies establish that touchscreen interactions leave informative motion responses. However, their targets, recording conditions and observation assumptions differ, so their reported accuracies are not directly comparable with isolated digit classification. Our study extends this line with an evaluation across participants and phone models, with drawing intervals assumed known.}

\noindent\textbf{\textcolor{black}{HuMIdb studies.}} \textcolor{black}{HuMIdb was introduced for bot detection~\cite{acien2020becaptcha,acien2021humidb} and subsequently used for risk-based authentication~\cite{papaioannou2022datasets} and for user verification with recurrent triplet networks and transformers~\cite{stragapede2022passive,senarath2023behaveformer,delgado2024swipeformer}. These applications exploit characteristics that distinguish people or separate human from automated interaction. Digit inference instead requires preserving information about the written symbol across variations in the person and the device. Consequently, representations developed for identity verification do not establish how well the same recordings support content inference.}

\noindent\textbf{\textcolor{black}{Temporal representations and contrastive learning.}} \textcolor{black}{MiniRocket extracts features from fixed convolutional kernels~\cite{dempster2021minirocket}, while patch transformers represent local temporal segments as tokens~\cite{nie2023patchtst}. SimCLR learns from augmented views of individual examples~\cite{chen2020simclr}, and time series methods extend contrastive learning through temporal and contextual objectives, including TS-TCC and the hierarchical representations of TS2Vec~\cite{eldele2021tstcc,yue2022ts2vec}. Supervised contrastive learning additionally uses shared class labels to define positive pairs~\cite{khosla2020supcon}, which provides a basis for pairing the same digit across people and phone models, and hard negative methods place more weight on similar representations from different classes~\cite{robinson2021hard}. However, augmentation benefits depend on the data and the model~\cite{iwana2021augmentation}, and cross-device positives do not necessarily establish generalization to unseen devices. These considerations motivate our comparison of temporal representations and contrastive objectives under explicit participant and phone model partitions.}

%
%
\section{Proposed Methodology}
\label{sec:method}

\textcolor{black}{The methodology has five stages, as shown in Figure~\ref{fig:overview}: cohort construction under an explicit observation model, signal conditioning, representation, classification, and evaluation under protocols that control which people and phone models are seen in training. Every fitted preprocessing operation uses the training partition only.}

\subsection{Problem formulation and observation model}
\label{sec:assumptions}

Each recording contains four motion streams $\mathcal{S}=\{\nacc,\lacc,\gyro,g\}$: the triaxial acceleration including gravity $\nacc(t)$, the triaxial linear acceleration $\lacc(t)$ estimated by Android, the triaxial angular velocity $\gyro(t)$, and gravity, stored by HuMIdb as a single value $g(t)$ that we treat as a scalar, since its recording means correlate with those of the accelerometer $x$ axis at 0.99. Each vector stream contributes its $x$, $y$ and $z$ components and their Euclidean norm, and gravity contributes $g$ and $|g|$, giving an input $\X\in\mathbb{R}^{C\times T}$ with $C=3\times4+2=14$ channels and $T=128$ samples. \textcolor{black}{For a recording with digit label $y\in\{0,\dots,9\}$, a classifier $f_\theta$ with parameters $\theta$ returns ten class scores, and the predicted digit is given by
\begin{equation}
\hat{y}=\argmax_{k\in\{0,\dots,9\}} f_\theta(\X)_k .
\label{eq:problem}
\end{equation}
Each recording also carries its participant $u$ and phone model $m$.} These identifiers define partitions and contrastive positives but never enter $f_\theta$, nor do touch values, absolute timestamps or task order. Four assumptions bound the claims.

\noindent\textbf{(A1) Unsupervised free task.} Participants used their own phones without supervision, while sitting, standing or walking~\cite{acien2021humidb}. Posture and support are not annotated, and a portrait interface does not imply a handheld phone. \textcolor{black}{Any statement about phones held in the hand or resting on a table is therefore indirect.}

\noindent\textbf{(A2) Oracle timing.} Touch events define the drawing interval, from the first valid contact to the last valid release. Sensor based segmentation is not evaluated. \textcolor{black}{The results therefore describe recognition once a drawing has been located, not the detection of drawings in a continuous stream.}

\noindent\textbf{(A3) One phone per participant.} User and physical device are confounded, so holding out phone model strings tests transfer across model families, not across the devices of one user.

\noindent\textbf{(A4) Fixed digit order.} Sessions request the digits 0 to 9 in order. \textcolor{black}{A deliberately unsafe control classifier that receives only the position of each recording within its session, as its start time offset and its rank, reaches 99.72\% test accuracy.} Chronology is excluded from all classifiers, but indirect task phase cues may remain.

\subsection{Signal conditioning}
\label{sec:conditioning}

\textcolor{black}{Signal conditioning maps the irregular native streams to a common grid, so that the same sample index refers to the same instant in every sensor.} After averaging repeated timestamps, each stream is cropped to the drawing interval given by touch, and the four streams are restricted to their common overlap $[t_s,t_e]$, from the latest first sample to the earliest last sample. \textcolor{black}{This avoids extrapolating a sensor that starts late or stops early.} Each primitive channel is interpolated onto a uniform grid at its median sample spacing, which defines its native rate, and resampled onto the shared grid \textcolor{black}{given by}:
\textcolor{black}{\begin{equation}
\tilde{t}_k=t_s+k\Delta,\qquad k=0,\dots,T-1,\qquad \Delta=\frac{t_e-t_s}{T-1}.
\label{eq:grid}
\end{equation}}%
When the output rate $1/\Delta$ is below the native rate, a sixth order Butterworth lowpass filter at $0.4/\Delta$ is applied forward and backward before resampling, suppressing aliasing without shifting the stroke phase. Norms are recomputed from the resampled axes, and channels are standardized with training statistics. The fixed length removes absolute duration, but the signal shape still reflects writing speed. \textcolor{black}{A drawing of 0.7~s, for example, is sampled at about 180~Hz, close to the median native rate of the database.}

\subsection{\textcolor{black}{Handcrafted features and classical classifiers}}

\textcolor{black}{The conventional baseline follows the usual practice of summarizing each channel with statistics that ignore the order of samples.} Each channel, interpolated to 50~Hz, is described by 29 statistics. Nineteen describe the amplitude distribution: moments, extremes, five percentiles, interquartile range, median absolute deviation, mean absolute value, root mean square (RMS), crest factor, zero crossing rate and lag one autocorrelation. Two describe the first difference, and eight describe the spectrum through its centroid, bandwidth, normalized entropy, dominant frequency and relative power in the 0 to 1, 1 to 3, 3 to 8 and 8 to 20~Hz bands. \textcolor{black}{Each vector sensor thus contributes 116 features and gravity 58, for 406 in total.} After median imputation and standardization, the features feed an RBF-SVM~\cite{cortes1995support}, a random forest~\cite{breiman2001random} and XGBoost~\cite{chen2016xgboost}, \textcolor{black}{which is reported as the strongest handcrafted reference.}

\subsection{\textcolor{black}{MiniRocket representation}}

MiniRocket~\cite{dempster2021minirocket} pools the responses of 84 fixed convolutional kernels over several dilations into 9,996 features, \textcolor{black}{which keep information about local patterns and their frequency along the recording without learning the kernels. PCA reduces them to 256 components for an RBF-SVM.} This representation uses an earlier tensor with independent sensor time axes and standardization within each recording. \textcolor{black}{Its rows in Table~\ref{tab:models} are therefore compared with a transformer trained on the same earlier tensor, and not with the final transformer.}

\subsection{\textcolor{black}{Sensor patch transformer}}

The proposed classifier tokenizes each sensor separately and fuses them through joint attention. For sensor $\nu$ with $C_\nu$ channels, let $\mathbf{x}_{\nu,i}\in\mathbb{R}^{C_\nu P}$ be the flattened patch of $P=8$ samples starting at sample $(i-1)S$, with stride $S=4$. A 1D convolution produces $N=1+(T-P)/S=31$ tokens per sensor\textcolor{black}{, expressed as}:
\begin{equation}
\mathbf{h}_{\nu,i}=\mathbf{W}_{\nu}\mathbf{x}_{\nu,i}+\mathbf{b}_{\nu}+\boldsymbol{\pi}_i+\mathbf{e}_\nu,\qquad i=1,\dots,N,
\label{eq:tokens}
\end{equation}
where $\mathbf{W}_{\nu}\in\mathbb{R}^{D\times C_\nu P}$ with $D=64$, $\boldsymbol{\pi}_i$ is a sinusoidal encoding of the patch position within the recording~\cite{vaswani2017attention}, and $\mathbf{e}_\nu$ is a learned sensor embedding. Separate projections keep the units and noise of each modality apart, and self attention over the $4N=124$ concatenated tokens fuses sensors and time at the patch level. \textcolor{black}{A patch of eight samples covers about 6\% of a drawing, so each token describes a short part of a stroke, and the positional encoding tells the encoder where in the drawing that part occurs.} Two encoder layers with four attention heads, a feed forward width of 256, GELU activations, dropout of 0.15 and layer normalization before each sublayer process the tokens, and a linear layer classifies their average. \textcolor{black}{The model has 108,426 parameters.}

\subsection{Training objectives and variants}

The transformer minimizes cross entropy with label smoothing~\cite{szegedy2016rethinking}\textcolor{black}{, which discourages overconfident predictions. The variants below test common extensions of this recipe, each changing one element.}

\vspace{1mm}
\noindent\textbf{Augmentation.} Each training input receives one integer shift $\delta\in\{-2,\dots,2\}$ shared by all sensors and Gaussian noise $\xi_{c,k}\sim\mathcal{N}(0,1)$ on the primitive channels\textcolor{black}{, so that the perturbed input is given by}:
\begin{equation}
\tilde{X}_{c,k}=X_{c,k'}+\eta\,\sigma_c\,\xi_{c,k},\qquad k'=\clamp(k+\delta;\,0,\,T-1),
\label{eq:augmentation}
\end{equation}
where $\clamp(x;a,b)=\min(\max(x,a),b)$, $\sigma_c$ is the training standard deviation of channel $c$ and $\eta=0.01$. The clamp holds edge values, so a stroke never wraps to its start, and norms are recomputed from the perturbed axes. \textcolor{black}{Sharing the shift across sensors keeps them aligned, and the small noise level keeps the digit recognizable.}

\vspace{1mm}
\noindent\textbf{Contrastive pretraining.} A projection head of width 64 with GELU maps the pooled embeddings of two augmented views to unit vectors $\mathbf{o}$. For an anchor $a$ with positives $\mathcal{P}_a$ and negatives $\mathcal{N}_a$ in the batch\textcolor{black}{, the contrastive loss is defined as}:
\begin{equation}
\mathcal{L}_a=-\frac{1}{|\mathcal{P}_a|}\sum_{p\in\mathcal{P}_a}\log\frac{e^{s_{ap}}}{\sum_{j\in\mathcal{P}_a}e^{s_{aj}}+\sum_{n\in\mathcal{N}_a}w_{an}e^{s_{an}}},
\label{eq:contrastive}
\end{equation}
with similarity $s_{aj}=\mathbf{o}_a^{\top}\mathbf{o}_j/\tau$ and temperature $\tau=0.1$. \textcolor{black}{The loss pulls each anchor towards its positives and pushes it away from its negatives, and the weights $w_{an}$ control how strongly each negative counts.} SimCLR~\cite{chen2020simclr} uses only the other view as positive. SupCon~\cite{khosla2020supcon} adds recordings of the same digit from other phone models, ignores those from the same phone model, and treats other digits as negatives. \textcolor{black}{Its positives therefore ask the encoder to map the same digit drawn on different phone models to nearby embeddings.} Plain training sets $w_{an}=1$, and the weighted variants, motivated by hard negative sampling~\cite{robinson2021hard}, use \textcolor{black}{the weights given by}:
\textcolor{black}{\begin{equation}
\begin{aligned}
w_{an}&=|\mathcal{N}_a|\,\frac{e^{\psi_{an}}}{\sum_{n'\in\mathcal{N}_a}e^{\psi_{an'}}},\\
\psi_{an}&=\beta\,\mathrm{sg}\big(\mathbf{o}_a^{\top}\mathbf{o}_n\big)+\big[\{y_a,y_n\}\in\mathcal{C}\big]\log\lambda,
\end{aligned}
\label{eq:weights}
\end{equation}}%
where $\mathrm{sg}$ stops the gradient, the Iverson bracket $[\cdot]$ equals one when its condition holds and zero otherwise, and $\mathcal{C}$ is a set of confusable digit pairs. \textcolor{black}{The factor $|\mathcal{N}_a|$ keeps the mean negative weight at one, so the weights only redistribute emphasis among the negatives.} The online variant ($\beta=2$, $\lambda=1$) raises the weight of negatives close to the anchor, and the pair variant ($\beta=0$, $\lambda=2$) doubles the relative weight of negatives whose digits form a pair in $\mathcal{C}=\{\{4,9\},\{6,8\},\{3,6\},\{1,7\}\}$. \textcolor{black}{These pairs were fixed in advance as hypotheses about visually similar shapes, not selected from validation or test confusion matrices.} After ten pretraining epochs, the head is discarded and the encoder is fine tuned \textcolor{black}{with the supervised objective and augmentation.}

\vspace{1mm}
\noindent\textbf{Derived channels and timing.} The jerk $d\lacc/dt$ and angular acceleration $d\gyro/dt$, estimated by central differences with time step $\Delta$, \textcolor{black}{add three axes each and give 20 input channels, and the trapezoidal integral of $\lacc$ adds three more for 23.} They are device frame signal features, not recovered velocity\textcolor{black}{, because orientation and drift are not corrected.} A timing control conditions the pooled embedding on the log duration\textcolor{black}{, to test whether the duration removed by the fixed grid carries digit information.}

%
%
\section{Experimental Evaluation}
\label{sec:evaluation}

\subsection{HuMIdb digit task and cohort}

HuMIdb contains 14 sensor streams from eight tasks recorded on the Android phones of the participants in up to five sessions at least one day apart~\cite{acien2021humidb}. \textcolor{black}{Its documentation reports more than 600 users and 179 devices. In the handwriting task, participants draw the digits 0 to 9 on the touchscreen with a finger.} Of the 599 participant folders and 24,840 digit recordings in our copy, 19,628 recordings from 481 participants and 165 phone model strings have a valid drawing interval and the four sensors. \textcolor{black}{Four database properties affect interpretation.} Sensor availability selects the population, as the accelerometer alone gives 24,556 recordings from 597 participants and adding the magnetometer leaves 19,251 from 474. Acquisition is irregular, as expected for heterogeneous phones~\cite{stisen2015smart}, with median native rate 196.51~Hz and gaps longer than 10\% of a sensor window in \textcolor{black}{237 of the 13,794 training recordings (1.72\%).} Orientation is always portrait, and the digit order gives the shortcut of (A4).

\subsection{Evaluation protocols}

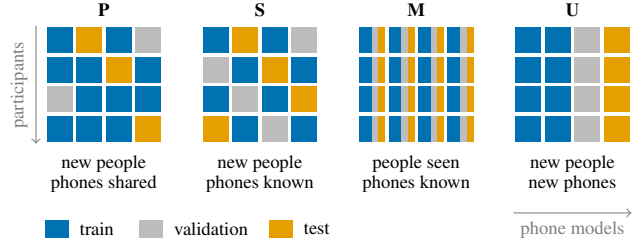
\begin{figure}[ht!]
\centering
\begin{tikzpicture}[x=0.39cm,y=0.39cm,font=\scriptsize]
  \newcommand{\panel}[4]{%
    \begin{scope}[shift={(#1,0)}]
      \foreach \x/\y/\c in {#3} {
        \fill[split\c] (\x+0.06,\y+0.06) rectangle (\x+0.94,\y+0.94);
      }
      \node[anchor=south] at (2,4.05) {\textbf{#2}};
      \node[anchor=north,align=center] at (2,-0.1) {#4};
    \end{scope}}
  \panel{0}{P}{0/3/tr,1/3/te,2/3/tr,3/3/va,0/2/tr,1/2/tr,2/2/te,3/2/tr,0/1/va,1/1/tr,2/1/tr,3/1/tr,0/0/tr,1/0/tr,2/0/tr,3/0/te}{new people\\phones shared}
  \panel{5.3}{S}{0/3/tr,1/3/te,2/3/tr,3/3/va,0/2/va,1/2/tr,2/2/te,3/2/tr,0/1/tr,1/1/va,2/1/tr,3/1/te,0/0/te,1/0/tr,2/0/va,3/0/tr}{new people\\phones known}
  \begin{scope}[shift={(10.6,0)}]
    \foreach \x in {0,...,3} { \foreach \y in {0,...,3} {
      \fill[splittr] (\x+0.06,\y+0.06) rectangle (\x+0.50,\y+0.94);
      \fill[splitva] (\x+0.50,\y+0.06) rectangle (\x+0.72,\y+0.94);
      \fill[splitte] (\x+0.72,\y+0.06) rectangle (\x+0.94,\y+0.94);
    } }
    \node[anchor=south] at (2,4.05) {\textbf{M}};
    \node[anchor=north,align=center] at (2,-0.1) {people seen\\phones known};
  \end{scope}
  \panel{15.9}{U}{0/3/tr,1/3/tr,2/3/va,3/3/te,0/2/tr,1/2/tr,2/2/va,3/2/te,0/1/tr,1/1/tr,2/1/va,3/1/te,0/0/tr,1/0/tr,2/0/va,3/0/te}{new people\\new phones}
  \draw[->,gray] (-0.3,4.0) -- (-0.3,0.0) node[midway,left,rotate=90,anchor=south,gray] {participants};
  \draw[->,gray] (15.9,-2.35) -- (19.9,-2.35) node[midway,below,gray] {phone models};
  \begin{scope}[shift={(0,-3.2)}]
    \fill[splittr] (0,0) rectangle (0.8,0.6); \node[anchor=west] at (0.9,0.3) {train};
    \fill[splitva] (3.2,0) rectangle (4.0,0.6); \node[anchor=west] at (4.1,0.3) {validation};
    \fill[splitte] (7.6,0) rectangle (8.4,0.6); \node[anchor=west] at (8.5,0.3) {test};
  \end{scope}
\end{tikzpicture}
\vspace{-2mm}
\caption{\textcolor{black}{Schematic of the four protocols. Columns are phone models and each cell is one participant, who uses a single phone (A3). In M, each participant contributes recordings to every partition. Proportions are illustrative.}}
\label{fig:protocols}
\end{figure}

\textcolor{black}{The four protocols, summarized in Table~\ref{tab:protocols},} differ in whether participants and phone models are shared between training and test\textcolor{black}{, as sketched in Figure~\ref{fig:protocols}.} \textbf{P} is the original participant holdout, fixed over all 599 folders before eligibility filtering, with 334, 72 and 75 disjoint participants for training, validation and test and overlapping phone models. \textbf{S}, \textbf{M} and \textbf{U} share 13,642 recordings from 333 participants, restricted to the 49 phone models used by at least three participants each.

\vspace{-2mm}
\begin{table}[ht!]
\centering
\caption{Protocols and transformer test accuracy (\%). Counts are recordings. 95\% bootstrap intervals over participants (P, S, M) or phone models (U). Last row: mean and SD over three seeds.}
\label{tab:protocols}
\footnotesize
\setlength{\tabcolsep}{3pt}
\begin{tabular}{@{}lccrrrcl@{}}
\toprule
 & \multicolumn{2}{c}{New at test} & & & & & \\
\cmidrule(lr){2-3}
Protocol & people & phones & Train & Test & People & Acc. & 95\% CI \\
\midrule
P & $\checkmark$ & $\times$ & 13,794 & 3,036 & 75 & 57.74 & [54.31, 61.18] \\
S & $\checkmark$ & $\times$ & 8,602 & 2,593 & 64 & 54.80 & [50.57, 58.69] \\
M & $\times$ & $\times$ & 8,602 & 2,593 & 323 & 56.69 & [54.15, 59.22] \\
U & $\checkmark$ & $\checkmark$ & 8,647 & 2,608 & 59 & 57.71 & [53.10, 61.59] \\
U, 3 seeds & $\checkmark$ & $\checkmark$ & 8,647 & 2,608 & 59 & 58.77 & SD 0.95 \\
\bottomrule
\end{tabular}
\end{table}

\textcolor{black}{This restriction makes it possible to hold out whole phone models while keeping several participants per model.} S partitions participants within each phone model, so test people are new but their phone models are represented in training. M randomly reassigns the S memberships to whole recordings within each phone model and digit stratum, preserving all stratum counts but introducing participant and session overlap, with 322 of the 323 test participants also represented in training. U assigns whole phone models (32, 8 and 9) to training, validation and test. \textcolor{black}{Comparing S and M isolates the effect of seeing the test people in training, and comparing S and U asks whether seeing their phone models matters.} S, M and U models are trained from scratch for 31 epochs, and the three seed U runs reuse one partition. Only the frozen P model was evaluated before any test result was inspected, and all other test results are retrospective.

\subsection{\textcolor{black}{Motion level analysis}}
\label{sec:motion_protocol}

\textcolor{black}{Because posture and phone support are unknown (A1), protocol U recordings are grouped by motion level, computed on the original timestamped samples of the drawing interval $[t_s,t_e]$ before any resampling. The motion is summarized by two time normalized measures, defined as:
\begin{equation}
\begin{aligned}
\rho_{\ell}^2&=\frac{1}{t_e-t_s}\int_{t_s}^{t_e}\big\|\lacc(t)-\bar{\lacc}\big\|^2dt,\\
\rho_{\omega}^2&=\frac{1}{t_e-t_s}\int_{t_s}^{t_e}\big\|\gyro(t)\big\|^2dt,
\end{aligned}
\label{eq:motion}
\end{equation}
where $\bar{\lacc}$ is the mean linear acceleration over the interval and the integrals use the trapezoidal rule on the native timestamps. Thus $\rho_\ell$ is the RMS variation of linear acceleration and $\rho_\omega$ is the RMS angular velocity. The combined score $\kappa=\max\{F_\ell(\rho_\ell),F_\omega(\rho_\omega)\}$ uses the empirical cumulative distribution functions $F_\ell$ and $F_\omega$ of the training recordings, which express each measure relative to the training data on a scale from zero to one. Taking the maximum means that a recording receives a low score only when both measures are small. Training tertiles define low ($\kappa<0.437$), medium ($0.437\leq\kappa<0.756$) and high ($\kappa\geq0.756$) motion, and the thresholds are applied unchanged to validation and test. These groups describe measured motion rather than labelled posture or support. To assess the contribution of low motion training data, we compare models trained on all recordings, without low motion recordings, and after random removal matched to the low motion removal in participant and digit counts, with three seeds per condition and the same test recordings.}

\subsection{Metrics and implementation details}

Accuracy is the primary metric, with macro-F1 and top-3 accuracy as secondary metrics. \textcolor{black}{Top-3 accuracy counts a recording as correct when the true digit is among the three most probable classes, which corresponds to an observer allowed three guesses.} The confidence intervals use a cluster bootstrap~\cite{efron1993bootstrap} with 2,000 resamples of participants, or of phone models for U, and paired differences resample whole clusters 5,000 times without correction for multiple comparisons. \textcolor{black}{Resampling whole clusters accounts for the correlation between recordings of the same person or phone model. The settings are listed in Table~\ref{tab:config}.} Early stopping on an inner split of 283 fitting and 51 stopping participants selected a budget of 31 epochs, after which a new model was trained on all 334 training participants. All S, M and U runs and seeds 2026 to 2028 reuse this budget.

\vspace{-2mm}
\begin{table}[ht!]
\centering
\caption{\textcolor{black}{Implementation settings.}}
\label{tab:config}
\footnotesize
\setlength{\tabcolsep}{3pt}
\begin{tabular}{@{}lp{0.72\columnwidth}@{}}
\toprule
\textcolor{black}{Component} & \textcolor{black}{Setting} \\
\midrule
\textcolor{black}{Conditioning} & \textcolor{black}{Shared overlap, median step grid, sixth order Butterworth at $0.4/\Delta$, $T=128$, $C=14$} \\
\textcolor{black}{Handcrafted} & \textcolor{black}{29 statistics per channel at 50~Hz, 406 features, median imputation, standardization} \\
\textcolor{black}{RBF-SVM} & \textcolor{black}{$C=10$, $\gamma=1/(d\,\mathrm{Var})$, Platt probabilities} \\
\textcolor{black}{Random forest} & \textcolor{black}{500 trees, $\sqrt{d}$ features per split, minimum leaf size 1} \\
\textcolor{black}{MiniRocket} & \textcolor{black}{9,996 features, at most 32 dilations per kernel, PCA to 256, RBF-SVM} \\
\textcolor{black}{Transformer} & \textcolor{black}{$P=8$, $S=4$, $D=64$, 2 layers, 4 heads, feed forward 256, GELU, dropout 0.15, 108,426 parameters} \\
\textcolor{black}{Optimization} & \textcolor{black}{AdamW~\cite{loshchilov2019adamw}, constant learning rate $3\times10^{-4}$, weight decay 0.01, batch 64, gradient clipping 1.0, FP16, label smoothing 0.05} \\
\textcolor{black}{Epochs} & \textcolor{black}{At most 40 with patience 8 on the inner stopping loss, 31 for the final refit} \\
\textcolor{black}{Augmentation} & \textcolor{black}{$\eta=0.01$, shared shift $|\delta|\le 2$, norms recomputed} \\
\textcolor{black}{Contrastive} & \textcolor{black}{10 pretraining epochs, projection of width 64 with GELU, $\tau=0.1$, $\lambda=2$ or $\beta=2$} \\
\textcolor{black}{Hardware} & \textcolor{black}{RTX 2080 SUPER with 8~GB, about 40~s to train and 0.1~s to classify the U test set} \\
\textcolor{black}{Software} & \textcolor{black}{Python 3.12, PyTorch 2.8.0, scikit-learn 1.7.2, sktime 0.40.1, XGBoost 3.0.5, SciPy 1.18.1} \\
\bottomrule
\end{tabular}
\end{table}

\vspace{-2mm}
%
%
\section{Results}
\label{sec:results}

\subsection{\textcolor{black}{Digit content across unseen participants}}

On protocol P, the frozen transformer classifies 1,753 of 3,036 test recordings correctly, giving 57.74\% accuracy, 57.77\% macro-F1 and 82.64\% top-3 accuracy against 10\% chance\textcolor{black}{, as shown in Table~\ref{tab:models}}.

\vspace{-2mm}
\begin{table}[ht!]
\centering
\caption{Accuracy (\%) on P, seed 2026 above, mean $\pm$ SD over three seeds below. $^{\dagger}$Independent sensor time axes, no antialiasing. Contrastive models are fine tuned with augmentation. \textcolor{black}{$^{\ast}$Selected on validation before the test set was opened, the only prospective test result. Other test values are retrospective.}}
\label{tab:models}
\footnotesize
\setlength{\tabcolsep}{10pt}
\begin{tabular}{@{}lrr@{}}
\toprule
Representation and training & Validation & Test \\
\midrule
Handcrafted, RBF-SVM & 39.67 & 39.72 \\
Handcrafted, random forest & 35.95 & 37.85 \\
MiniRocket, PCA 256, RBF-SVM$^{\dagger}$ & 54.04 & 52.64 \\
Transformer$^{\dagger}$ & 56.75 & 55.67 \\
\quad without temporal positions$^{\dagger}$ & 35.67 & 35.97 \\
\midrule
Transformer (selected)\textcolor{black}{$^{\ast}$} & 60.29 & 57.74 \\
\quad with noise and shifts & 59.65 & 57.67 \\
\quad SimCLR pretraining & 57.65 & 56.49 \\
\quad SupCon across phone models & 59.65 & 57.58 \\
\quad SupCon, weighted digit pairs & 59.83 & 56.69 \\
\quad SupCon, online hard negatives & 58.97 & 57.61 \\
\midrule
Transformer, three seeds & 60.91 $\pm$ 1.26 & 57.75 $\pm$ 1.14 \\
\quad with jerk and angular acceleration & 60.85 $\pm$ 0.20 & 58.48 $\pm$ 0.54 \\
\quad with derivatives and integrated $\lacc$ & 60.50 $\pm$ 1.14 & 58.54 $\pm$ 0.56 \\
\quad with log duration context & 61.25 $\pm$ 1.90 & 58.54 $\pm$ 1.19 \\
\bottomrule
\end{tabular}
\end{table}

\begin{figure}[ht!]
\centering
\includegraphics[width=0.98\columnwidth]{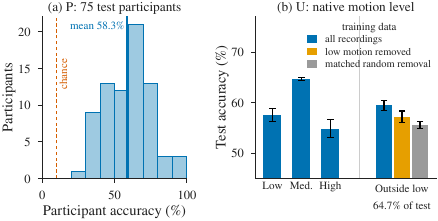}
\vspace{-2mm}
\caption{(a) Test accuracy of the 75 P participants. (b) U accuracy by motion level (training on all recordings) and outside the low group (64.7\% of test) for three training sets, mean and SD over three seeds.}
\label{fig:participants}
\end{figure}

\begin{figure}[ht!]
\centering
\includegraphics[width=0.98\columnwidth]{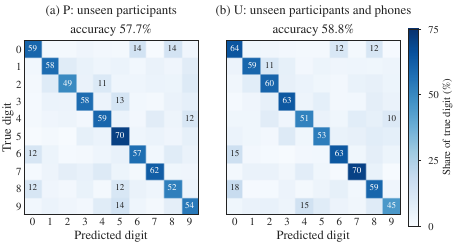}
\vspace{-2mm}
\caption{Test confusion matrices normalized by true digit (\%). (a) P, frozen transformer, 3,036 recordings, 75 unseen participants. (b) U, three seeds pooled, 2,608 recordings, 59 unseen participants, nine unseen phone models. Off diagonal values below 10\% omitted.}
\label{fig:confusions}
\end{figure}

 \textcolor{black}{The handcrafted RBF-SVM and random forest reach 39.72\% and 37.85\%, the strongest handcrafted classifier, XGBoost, reaches 42.23\%, and MiniRocket reaches 52.64\%.} Removing positional encodings lowers transformer accuracy from 55.67\% to 35.97\%. A few participants do not drive the result, as shown in Figure~\ref{fig:participants}a\textcolor{black}{. Participant accuracy ranges from 22.5\% to 100\% with a mean of 58.3\%, all 75 participants exceed 20\%, 69.3\% reach at least 50\%, and accuracy ranges only from 56.11\% to 60.56\% across sessions.} Errors concentrate on the digits 0, 6 and 8 and on the pair 4 and 9, as shown in Figure~\ref{fig:confusions}a\textcolor{black}{, and the per digit F1 ranges from 50.7\% for the digit 9 to 63.9\% for the digit 7.}

\vspace{-2mm}

\noindent\textbf{Observations.} The 15.5 point gap to the strongest handcrafted classifier and the collapse without positions indicate that much of the digit information lies in the order of motion within a stroke, which summary statistics discard. \textcolor{black}{MiniRocket, whose kernels respond to local temporal patterns, recovers most of this gap.} The confused digits share parts of their paths, such as the loops of 0, 6 and 8, consistent with a response to stroke shape, although shape is not isolated from other cues.

\vspace{-2mm}

\subsection{\textcolor{black}{Sensor contribution}}

In the fixed cohort of 19,251 recordings with five sensors, the handcrafted RBF-SVM reaches 32.99\% with the gyroscope alone, 25.86\% with the accelerometer and 24.76\% with linear acceleration. The three motion sensors give 38.23\% together, 40.17\% with gravity and 37.49\% with the magnetometer instead.

\noindent\textbf{Observations.} Finger movement rotates the phone slightly, and the gyroscope measures this rotation directly as angular velocity, whereas the accelerometer is dominated by gravity and orientation. The magnetometer follows the environmental field and heading rather than the digit, as reported for selfie capture motion~\cite{rantahalvari2026selfie}, so more sensors need not add digit information. \textcolor{black}{These comparisons use handcrafted features, and the ranking may differ for learned representations.}

\subsection{\textcolor{black}{Unseen phone models and participant overlap}}

The removal of phone models does not reduce the accuracy in our partitions\textcolor{black}{, as shown in Table~\ref{tab:protocols}}. U reaches 57.71\% and 58.77\% over three seeds, against 54.80\% for new people on known phone models in S. The populations differ, so unseen phones are not shown to be easier, but accuracy does not require the phone model of a test user in training. \textcolor{black}{On U, the top-3 accuracy is 84.39\%, and accuracy across the nine held out phone models ranges from 22.2\% to 81.4\%. Both extremes come from models with three participants each, while the most represented model, with 940 recordings from 20 participants, reaches 56.7\%.} Mixing M recordings raises the accuracy only to 56.69\%, 1.89 points above S, despite the participant overlap.

\noindent\textbf{Observations.} P and U share the same confusion structure, as depicted in Figure~\ref{fig:confusions}. Digit motion patterns thus appear largely shared across people and phones, unlike the user specific motion that authentication exploits~\cite{sitova2016hmog}. \textcolor{black}{The small gain from participant overlap suggests that the models do not rely strongly on memorizing individual writers.} This evidence is based on one U partition with nine test models (A3)\textcolor{black}{, and the wide spread across phone models with few participants shows that single model estimates are uncertain.}

\subsection{Training extensions}

The tested extensions give no consistent top-1 accuracy gains\textcolor{black}{, as reported in Table~\ref{tab:models}}. On validation, SupCon differs from the baseline by $-0.64$ points (paired 95\% CI [$-2.48$, 1.30]). Weighting confusable pairs reduces the errors within them from 171 to 165 but lowers accuracy to 56.69\%. Derived channels change test accuracy by $+0.72$ points on P (CI [$-0.41$, 1.86]) and by $-0.64$ points on U (CI [$-2.23$, 0.37]), whereas a log duration context adds $+0.79$ points on P (CI [0.15, 1.46]), with a validation interval that includes zero.

\noindent\textbf{Observations.} With 13,794 labelled recordings, the digit labels already define the classes SupCon separates, so contrastive pretraining adds little. The augmentation is deliberately conservative, since stronger warping or rotation would alter the timing and rotation that carry the digit. Derivatives are local filters the patch projection can learn, and the duration context restores timing lost to the fixed length grid.

\subsection{Motion level and unknown phone support}

The low motion group is not motionless (median gyroscope RMS 0.127~rad/s), and only 8.9\% of its test recordings indicate a near horizontal screen, so it does not label phones on a table. Trained on all recordings, the model reaches 57.54\% on low, 64.65\% on medium and 54.81\% on high motion, as depicted in Figure~\ref{fig:participants}b. Removing low motion recordings from training lowers accuracy on the same test recordings outside the low group from 59.44\% to 57.17\% ($-2.27$ points, CI [$-3.58$, $-1.43$] over phone models), and matched random removal gives 55.57\%.

\noindent\textbf{Observations.} The non monotonic pattern suggests a balance between weak reaction forces at low motion and grip changes or body movement at high motion. Low measured motion does not prevent inference, and low motion recordings also help the model on other recordings. Motion level partly tracks writing speed, since low motion drawings last longer (median 0.84~s against 0.68~s).

\subsection{\textcolor{black}{Implications for privacy}}

\textcolor{black}{For a new person, the correct digit is among the three most probable guesses in 82.64\% of recordings, and knowledge of the phone model is not required. For a short numeric code, this could narrow the search considerably, although the combination of several digits was not evaluated. The ablations show that temporal order and rotation carry most of the information, which motivates investigating perturbations of the timing and rotation content of motion streams as possible defences. Their effectiveness, and their cost for legitimate applications, were not evaluated. A practical attacker would also need to detect the drawing interval from motion alone (A2).}

%
%
\section{Implications and Conclusion}
\label{sec:conclusion}

We present a characterization of handwritten digit predictability from smartphone motion, comparing three representation families under evaluation protocols that control participant and phone model overlap. Assuming known drawing intervals, a compact transformer classifies ten digits with 57.74\% accuracy on unseen participants and 58.77$\pm$0.95\% across three training seeds on unseen participants using unseen phone models. Low motion recordings remain informative, and accuracy does not increase monotonically with motion level. The tested contrastive pretraining, augmentation and derived channels provide no consistent gains. These results establish predictability across participants and phone models within the evaluated setting, suggesting that the privacy implications of motion collection warrant consideration beyond identity and behavioural inference. The findings remain conditional on known drawing intervals, one phone per participant, a single phone model holdout partition and retrospective comparisons. Fixed digit order is the principal threat to validity because indirect acquisition cues may contribute to classification even when explicit chronology is excluded. Consequently, the results do not establish how much predictive information arises specifically from handwriting dynamics or demonstrate an operational attack. Future evaluations should randomize digit order, annotate posture and device support, and incorporate sensor based detection of drawing intervals. The temporal and sensor ablations also motivate investigating perturbations of temporal and rotational information as potential defenses, whose effectiveness and impact on legitimate applications remain to be evaluated.

\section*{Acknowledgment}
The research was supported by the Research Council of Finland Smart Video Sensorization for Secure Healthcare Monitoring (SViSenS) project (370277), the Interreg Aurora ResilientEdge project (20373282), Business Finland through the HBIAS project (3092/31/2025) and the Profi7 Hybrid Intelligence programme (352788). AI assisted tools were used for language editing and coding, and Elicit as a literature discovery aid. The authors made all inclusion, extraction, synthesis, interpretation and writing decisions.

\section*{Compliance with Ethical Standards}
This retrospective study used human subject data from HuMIdb~\cite{acien2021humidb}, collected with participant consent, anonymised under the GDPR and obtained under the license agreement of the database providers. No additional ethical approval was required for this secondary analysis.

\bibliographystyle{IEEEbib}
\bibliography{references}

\end{document}